%% file: main.tex
\documentclass[final]{cvpr}

\usepackage{times}
\usepackage{epsfig}
\usepackage{graphicx}
\usepackage{amsmath}
\usepackage{amssymb}
\usepackage{comment}
\usepackage{color}
\usepackage{graphicx}
\usepackage{graphics}
\usepackage{subcaption}
\usepackage{makecell}

\usepackage{mathtools, cuted}
\usepackage{lipsum, color}

\usepackage[pagebackref=true,breaklinks=true,colorlinks,bookmarks=false]{hyperref}

\def \W{\mathbf W}
\def \WW{{\boldsymbol{\mathcal W}}}
\def \X{\mathbf X}

\def \L{\mathcal L}
\def \bbeta{{\boldsymbol \beta}}
\def \mmu{{\boldsymbol \mu}}
\input{math_commands.tex}

\def\cvprPaperID{4} 
\def\confYear{CVPR 2021}

\begin{document}

\title{Adaptive Binary-Ternary Quantization}



\author{Ryan Razani, Gr\'egoire Morin,  Eyy\"ub Sari, and Vahid Partovi Nia \\
Huawei Noah's Ark Lab \\
{\tt \small ryan.razani@huawei.com}
}

\maketitle

\thispagestyle{empty}

\begin{abstract}
   Neural network models are resource hungry. It is difficult to deploy such deep networks on devices with limited resources, like smart wearables, cellphones, drones, and autonomous vehicles. Low bit quantization such as binary and ternary quantization is a common approach  to alleviate this resource requirements. Ternary quantization provides a more flexible model and outperforms binary quantization in terms of accuracy, however doubles the memory footprint and increases the computational cost. Contrary to these approaches, mixed quantized models allow a trade-off between accuracy and memory footprint. In such models, quantization depth is often chosen manually, or is tuned using a separate optimization routine. The latter requires training a quantized network multiple times. Here, we propose an adaptive combination of binary and ternary quantization, namely Smart Quantization (SQ), in which the quantization depth is modified directly via a regularization function, so that the model is trained only once. Our experimental results show that the proposed method adapts quantization depth successfully while keeping the model accuracy high on MNIST and CIFAR10 benchmarks.
\end{abstract}

\section{Introduction}

\noindent Deep Neural Network (DNN) models have achieved tremendous attraction due to their success on a wide variety of tasks, including computer vision, automatic speech recognition, natural language processing, and reinforcement learning \cite{DL_book}. More specifically, in computer vision DNN have led to a series of breakthrough for image classification \cite{Alexnet}, \cite{ImageRecog}, \cite{DeeperConv}, and object detection \cite{Yolo}, \cite{SSD}, \cite{Faster_RCNN}. DNN models are computationally intensive and require large memory to store the model parameters. Computation and storage resource requirement becomes the main obstacle to deploy such models in many edge devices due to lack of memory, computation power, energy, etc.  This motivated the researchers to develop compression techniques to reduce the cost for such models.

Recently, several techniques have been introduced in the literature to solve the storage and computational limitations of the edge devices. Among them, quantization methods focus on representing the weights of a neural network in lower precision than the usual $32$-bits float representation, saving on the memory footprint of the model. Binary quantization \cite{courbariaux2015binaryconnect}, \cite{hubara2016binarized}, \cite{XNORNet}, \cite{DoReFaNet}, \cite{ABCNet}, \cite{darabi2018regularized} represent weights with 1 bit precision and ternary quantization \cite{TerConnect}, \cite{TWN}, \cite{TTQ} with 2 bits precision. While the latter frameworks lead to significant memory reduction compared to their full precision counterpart, they are constrained to quantize the model with 1 bit or 2 bits, on demand. We relax this constraint, and present Smart Quantization (SQ) that allows adapting layers to 1 bit and 2 bits while training the network. Consequently, this approach automatically quantizes weights into binary or ternary depending upon a trainable control parameter. We show that this approach leads to mixed bit precision models that beats ternary networks both in terms of accuracy and memory consumption. Here we only focus on quantizing layers because it is more feasible to implement layer-wise quantization at inference time after training. However, this method can be also adapted for mixed precision training of sub-network, block, filter, or weight.

\section{Related Work}
There are two main components in DNN models, namely, weights and activations. These two components are usually computed in full precision, i.e. floating point 32-bits. This work focuses on quantizing the weights of the network, i.e. generalizing  BinaryConnect (BC) \cite{courbariaux2015binaryconnect} and Ternary Weight Network (TWN) \cite{TWN} toward automatic 1 or 2 bits mixed-precision using a single training algorithm. 

In BC \cite{courbariaux2015binaryconnect}, the real value weights, $w$, are binarized to $w^{b}\in \{-1,+1\}$ during the forward pass. To map a full precision weight to a binary weight, the deterministic $sign$ function is used,
\begin{equation}
{w}^{b} = \text{sign}(w) = 
\begin{cases}
	+1  &	{w} \geq 0,\\
	-1  & {w} < 0.
\end{cases}
\end{equation}
The derivative of the $sign$ function is zero on $\mathbb{R} \setminus \{0\}$. The sign function has zero gradient which freezes weight updates during back-propagation. To bypass this problem, BC \cite{courbariaux2015binaryconnect} uses a clipped straight-through estimator 
\begin{equation}
\frac{ \partial \L}{\partial w} = \frac{\partial \L}{\partial w^b}\textbf{1}_{|w| \leq 1}(w)
\label{eq:straight-through}
\end{equation} 
where $\L$ is the loss function and $\textbf{1}_A(.)$ is the indicator function on the set $A$. In other words \eqref{eq:straight-through} approximates the $\text{sign}$ function by the linear function $f(x) = x$ within $[-1, +1]$ and a constant elsewhere. During back propagation, the weights are updated only within $[-1,+1]$. The binarized weights are updated with their corresponding full precision gradients. XNOR-Net \cite{XNORNet} adds a scaling factor to reduce the gap between binary and full-precision model's accuracy, defining Binary Weight Network (BWN). The real value weights $\W$ in each layer are quantized as $\mu \times \{ -1, +1 \}$ where $\mu = \mathbb{E}\big[|\W|\big] \in \mathbb{R}$. DoReFa-Net \cite{DoReFaNet} generalizes the latter work and approximates the full precision weights with more than one bit, while ABCNet \cite{ABCNet} approximates weights with a linear combination of multiple binary weight bases. 

\subsection{Ternary weight networks }

\noindent Ternary Weight Network (TWN) \cite{TWN} is a neural network with weights constrained to $\{-1,0,+1\}$. The weight resolution is reduced from 32 bits to 2 bits, replacing full precision  weights with ternary weights. TWN aims to fill the gap between full precision and binary precision weight. Compared to binary weight networks, ternary weight networks are more expressive. In  3$\times$3 weight filter in a convolutional neural network, there are $2^{3 \times 3}=512$ possible variations with binary precision and $3^{3 \times 3}=19683$ with ternary precision. 

TWN \cite{TWN} finds the closest ternary weights matrix $\W^t$ to its corresponding real value weight matrix $\W$ using
\begin{equation}
\left\{
  \begin{array}{lr}
    \hat{\mu}, \hat{\W}^{t}=  \argmin \limits_{\mu, \W^{t}}  \Vert \textbf{W}-\mu {\textbf{W}}^{t} \Vert^{2}_{2},\\
    s.t. \> \> \mu \geq 0, \> \> {w}^{t}_{ij} \in \{-1,0,1\}, \> \> i,j=1,2,...,n.
  \end{array}
\right.
\label{eq:ternary_opt}
\end{equation}
The ternary weight $\W^{t}$ is achieved by applying a symmetric threshold $\Delta$
\begin{equation}
{\W}^{t} = 
  \begin{cases}
    +1 & 	{w}_{ij} > \Delta,\\
    0 & 	 |{w}_{ij}| \leq \Delta,\\
    -1 &  {w}_{ij} < -\Delta.
  \end{cases}
\label{eq:thr_function}
\end{equation}
One may adopt a weight-dependant threshold $\Delta$ and a scaling factor $\mu$ that approximately solves \eqref{eq:ternary_opt}.   Similar to BC and BWN schemes; ternary-value weights are only used for the forward pass and back propagation, but not for the parameter updates. At inference, the scaling factor can be folded with the input $\X$ as,
\begin{equation}
\X\odot \W  \approx \X \odot (\mu \W^{t})= (\mu  \X) \odot \W^{t},
\end{equation}
where $\odot$ denotes the convolution.

Trained Ternary Quantization (TTQ) \cite{TTQ} proposes a more general ternary method which reduces the precision of weights in neural network to ternary values. However, TTQ quantizes the weights to asymmetric values $\{-\mu_1, 0, +\mu_2\}$ using two full-precision scaling coefficients $\mu_1$ and $\mu_2$ for each layer of neural network.  Consequently, the method achieves better accuracy as opposed to TWN.

Note that $\mu$ is regarded as a scaling factor when the network is quantized,  during training, within a certain structure (e.g., per layer, per filter, per channel). Varying $\mu$ per weight may end up canceling the ternary simplification which requires the same hardware as a full precision network.
However, restricting $\mu$ in the form of $2^n$ simplifies multiplication to shift operation to the right or left depending on the sign of the integer value $n \in \mathbb{Z}$, see \cite{wu2018shift}.

Our method provides a compromise between BC and TWN and trains weights with a single trainable scaling factor $\mu$. Weights shift between ternary $\{-\mu, 0, +\mu\} $ and binary $\{-\mu, +\mu\}$. This provides a single algorithm for 1 or 2 bits mixed precision.

\subsection{Regularization}
Regularization technique is essential to prevent over-fitting problem and to obtain robust generalization for unseen data. Standard regularization functions, such as $L_2$ or $L_1$ encourage weights to be concentrated about the origin. However, in case of binary network it is more appropriate to have a regularization function to encourage the weights about $\mu \times\{-1,+1\}$, with a scaling factor $\mu >0$ \cite{nia2018binary} such as ,
\begin{equation}
R_1(w, \mu) =  \big\vert \vert w \vert-\mu \big\vert,
\label{eq:binquant}
\end{equation}

\begin{figure}
	\centering
	\begin{subfigure}[h!]{0.45\linewidth}
		\includegraphics[width=\textwidth]{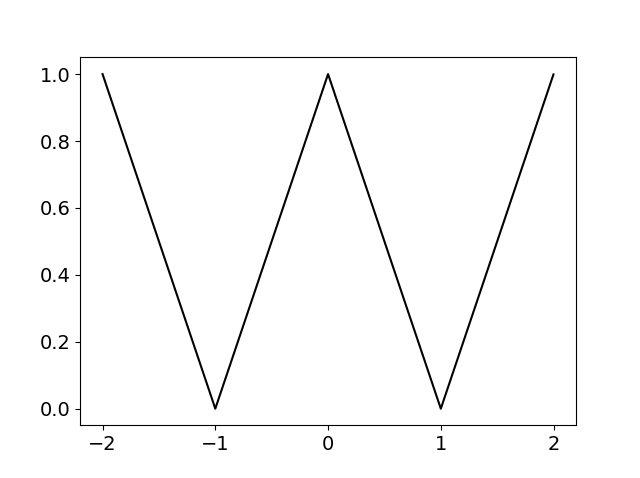}
		\caption{$R_1(w,\mu = 1)$}
		\label{fig:1bit_reg}
	\end{subfigure}
	~ 
	\begin{subfigure}[h!]{0.45\linewidth}
		\includegraphics[width=\textwidth]{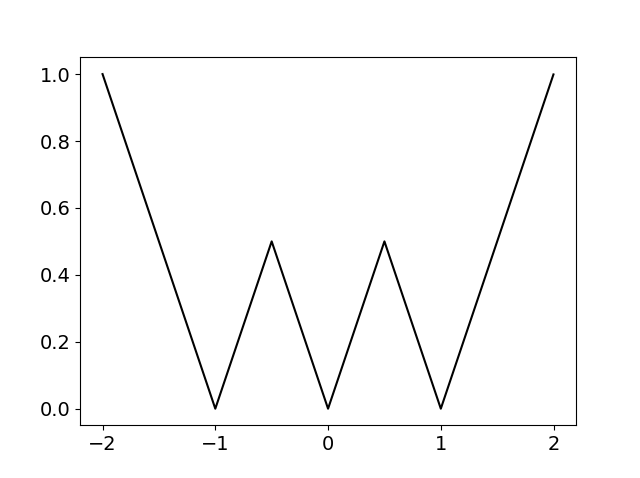}
		\caption{$R_2(w,\mu = 1)$}
		\label{fig:2bit_reg}
	\end{subfigure}
	\caption{Binary and ternary regularizers; $R_1$ encourages binary weights, with minimums at $\{-\mu,+\mu\}$, and $R_2$ encourages ternary weights, with minimums at $\{-\mu,0,+\mu\}$.}
	\label{fig:1_2bit_reg}
\end{figure}

\noindent A straightforward generalization for ternary quantization can be expressed as,
\begin{equation}
R_2(w, \mu) = \bigg\vert \big\vert \vert w \vert-{ \frac{\mu}{2} } \big\vert - \ { \frac{\mu}{2} } \bigg\vert.
\label{eq:terquant}
\end{equation}
Regularizer \eqref{eq:binquant} encourages weights about $\{-\mu,+\mu\}$, and \eqref{eq:terquant} about $\{-\mu , 0, +\mu\}$. The two functions are depicted in Figure~\ref{fig:1_2bit_reg}. These regularization functions are only useful when the quantization depth is set before training starts. We propose a more flexible version to smoothly move between these two functions using a shape parameter $\beta$.

\section{Adaptive Quantization}
\begin{figure*}[htbp!]
	\centering
	\begin{subfigure}[t]{0.32\linewidth} 
		\includegraphics[width=\textwidth]{files/2bit_reg.png}
		\caption{$R(w,\beta = \frac{\pi}{4})$}
		\label{fig:stq_reg1}
	\end{subfigure}
	\begin{subfigure}[t]{0.32\linewidth} 
		\includegraphics[width=\textwidth]{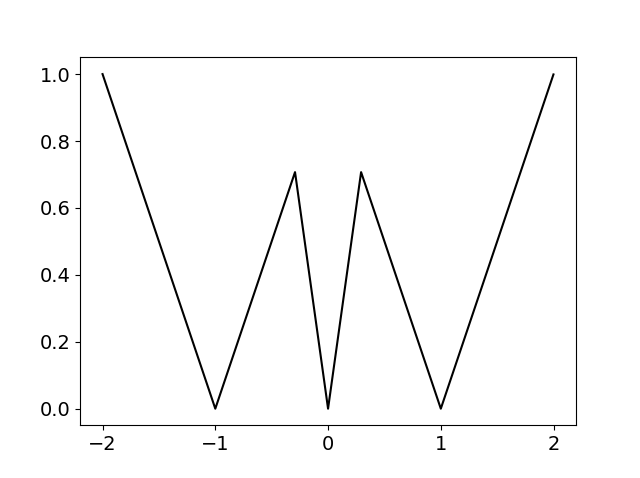}
		\caption{$R(w, \beta = \frac{3 \pi}{8})$}
		\label{fig:stq_reg2}
	\end{subfigure}
	\begin{subfigure}[t]{0.32\linewidth} 
		\includegraphics[width=\textwidth]{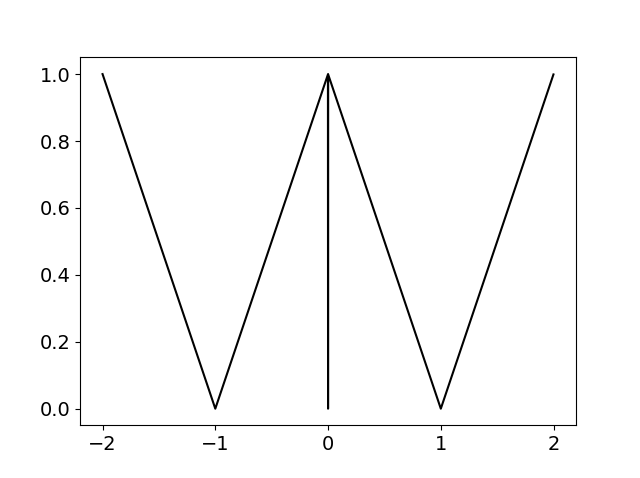}
		\caption{$R(w, \beta = 1.5707)$}
		\label{fig:stq_reg3}
	\end{subfigure}
	\caption{Adaptive regularization function. When $\beta \to {\pi \over 2}$ the regularization function switches from  ternary to binary.}
	\label{fig:stq_reg}
\end{figure*}

Here we propose a generalized adaptive regularization function that switches between binary regularization of \eqref{eq:binquant} and ternary regularization of \eqref{eq:terquant} 
\begin{equation}
\min \bigg( \big\vert |w|+ \mu\big\vert^p, \big\vert |w|- \mu\big\vert^p, ~ \tan(\beta)|w|^p \bigg),
\label{eq:binter}
\end{equation}
in which $\mu$ is a trainable scaling factor, $p$ is the order coefficient denoting the type of regularization function, and $\beta \in ({\frac{\pi}{4}} , {\frac{\pi}{2}})$ controls the transition between \eqref{eq:binquant} and \eqref{eq:terquant}. As a special case $\beta \to {\frac{\pi}{2}} $ converges to the binary regularizer \eqref{eq:binquant}  and $ \beta \to {\frac{\pi}{4}}$ coincide with the ternary regularizer \eqref{eq:terquant}, depicted in Figure~\ref{fig:stq_reg}. A large value of $\tan(\beta)$ repels estimated weights away from zero thus yielding binary quantization, and a small value of $\tan(\beta)$ encourages zero weights.  The shape parameter $\beta$ controls the quantization depth. Quantization depth changes per layer; therefore we let $\beta$ vary per layer. We recommend to regularize $\beta$ about $\frac{\pi}{2}$ i.e. preferring binary quantization apriori by adding $|\cot(\beta)|$ to the objective function.

For a single filter $\W$ the regularization function can be expressed by a sum over its elements on row $i$ and column $j$ as,
\begin{align} \label{eqn_einstein}
\begin{split}
    R(\W,\mu, \beta) = {}& \sum_{i=1}^I\sum_{j=1}^J \min \bigg( \big\vert |w_{ij}|+ \mu\big\vert^p, \big\vert |w_{ij}|- \mu\big\vert^p , \\  
    &  ~ \tan(\beta)|w_{ij}|^p \bigg) + \gamma |\cot(\beta)|,
\end{split}
\end{align}

\noindent where $\gamma$ controls the proportion of binary to ternary layers, i.e., large values of $\gamma$ contributes to encouraging the layers to form binary values. 
In each layer, weights are pushed to binary or ternary values, depending on the trained value of the corresponding $\beta$. Here we only focus on such regularizer that is constructed using $p=1$, as the accuracy does not change significantly by varying the value of $p$. 

The introduced regularization function is added to the empirical loss function $L(.)$. The objective function is optimized on set of parameter weights $\WW$, set of scaling factors $\mmu$, and set of shape parameters $\bbeta$  using back propagation  
\begin{align}
  \L(\WW,\mmu,\bbeta) =  L(\WW) +  \sum_{l=1}^L \lambda_l \sum_{k=1}^{K}  R(\W_{kl}, \mu_{kl},\beta_l),
  \label{eq:obj_smart}
\end{align}
where $k$ indexes the channel in a convolution network, and $l$ indexes the layer. One may use a different regularization constant $\lambda_l$ for each layer to keep the impact of the regularization term balanced across layers, indeed different layers may involve different number of parameters. We set $\lambda_l = {\lambda \over \# \W_{l} }$ where $\lambda$ is a constant, and $ \# \W_{l}$ is the number of weights in layer $l$.

We propose to use the same threshold-based function of TWN \cite{TWN}  \eqref{eq:thr_function}, but with a fixed threshold $\Delta_l$ per layer $l$. Note that \cite{TWN} proposes a weight-dependant threshold. We enforce weights to only accumulate about $\{ -\mu, + \mu \}$ for large $\beta_l$. One may set $\Delta_l$ to have the same balanced weights in $\{-\mu, 0, +\mu\}$ at initialization for all layers and let the weights evolve during training. Formally, if $\sigma_l$ is the standard deviation of the initial Gaussian weights in layer $l$, we propose $\Delta_l = 0.2 \times \sigma_l. $ The probability that a single weight lies in the range $[-\Delta_l,\Delta_l]$ is $ \approx 0.16$. All the weights falling in this range will be quantized as zeros after applying the threshold function. 

Weights are naturally pushed to binary or ternary values depending on $\beta_l$ during training. Eventually, a threshold $\delta$ close to ${\pi \over 2}\approx 1.57$ defines the final quantization depth for each layer. 

\vspace{-2mm}
\begin{equation*}
\text{Final quantization depth of layer }l :
  \begin{cases}
    \text{Binary}  & 	\beta_l   \geq  \delta,\\
    \text{Ternary} & 	\beta_l     <   \delta
  \end{cases}
\end{equation*}
\vspace{-2mm}

\section{Experiments}
We run experiments on two common image classification tasks MNIST \cite{MNIST} and CIFAR10 \cite{krizhevsky2009learning} datasets. We compare our method, Smart Quantization (SQ), with BinaryConnect (BC) of \cite{courbariaux2015binaryconnect}, Binary Weight Networks (BWN) of \cite{XNORNet}, Ternary Weights Network (TWN) of \cite{TWN} and also with a Full Precision network (FP). The quality of the compression is measured only in terms of memory, it is difficult to compare mixed precision models, with binary and ternary, in terms of consumed energy as their fair comparison requires specific hardware design. Let $n_l$ be the quantization depth for the layer $l$ and $\# \W_{l}$ the number of weights in layer $l$, therefore the  compression ratio is derived as, $ \frac{\sum_{l=1}^L \# \W_{l} \times 32}{\sum_{l=1}^L \# \W_{l} \times n_l}.$
The compression ratio for a binary network is $32$, for a ternary network $16$, and our approach falls in between.

\begin{figure}[h]
	\centering
	\begin{subfigure}[t]{0.47\linewidth}
		\includegraphics[width=\textwidth]{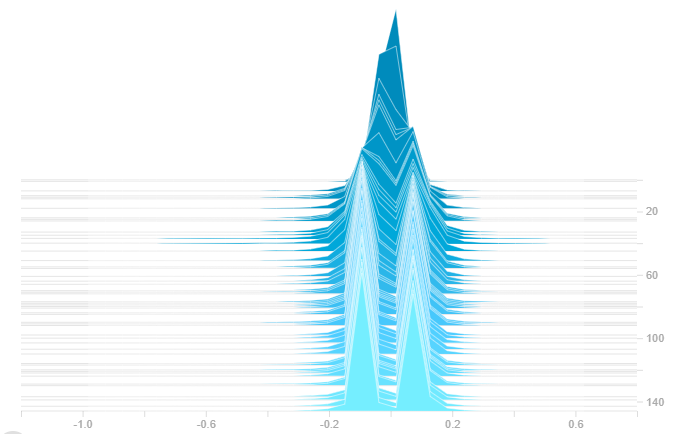}
		\caption{$\gamma = 1\times 10^{-1}$, $\beta = 1.57$}
		\label{fig:bigger_gamma}
	\end{subfigure}
	~ 
	\begin{subfigure}[t]{0.47\linewidth}
		\includegraphics[width=\textwidth]{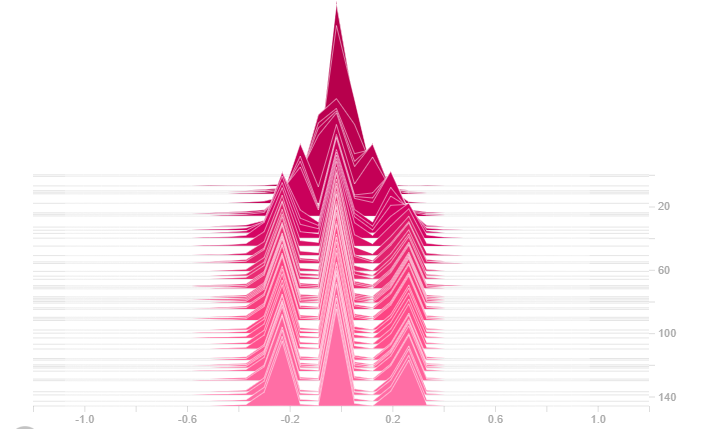}
		\caption{$\gamma = 1\times 10^{-5}$, $\beta = 0.79$}
		\label{fig:smaller_gamma}
	\end{subfigure}
	\caption{Effect of $\gamma$ on the weights distribution of a layer while training. a) binray and b) ternary weight distribution, respectively.}
	\label{fig:gamma_effect}
\end{figure}

Our SQ network generalizes binary and ternary regularization in a single regularization function. We present how to control the proportion of binary and ternary layers using $\gamma$ in \eqref{eq:binter}. Figure~\ref{fig:gamma_effect} clarifies the effect of $\gamma$ on the weight distribution. When $\gamma$ is large, $\beta$ is encouraged towards $\pi \over 2$ which corresponds to binary quantization. Thus, weights are pushed about $\{-\mu,+\mu\}$ and $0$ is removed from the trained values, see Figure~\ref{fig:bigger_gamma}. On the contrary, when $\gamma$ is small, $\beta$ tends to $\pi \over 4$ and the weights started including 0 in their values, see Figure~\ref{fig:smaller_gamma}. 
In our experiment, the parameters $\WW$, $\mmu$ and $\bbeta$ are trainable; $p$, $\delta_l$, $\Delta_l$, $\lambda_l$, and $\gamma$ are tuning parameters.

\subsection{MNIST}
MNIST is an image classification benchmark dataset with  28 $\times$ 28 gray-scale images representing digits ranging from $0$ to $9$. The dataset is split into 60k training images and 10k testing images. We used the LeNet-5 \cite{MNIST} architecture consisting of 5 layers, 2 convolution followed by maxpooling, stacked with two fully connected layers and a softmax layer at the end. We train the network for $60$ epochs using Adam optimizer. We used the initial learning rate of $0.01$, but divided by $10$ in epoch $15$ and $30$ to stabilize training. The batch size is set to $64$ with $L_2$ weight decay constant  $10^{-4}$ only for BC, BWC and TWN. The full precision LeNet-5 is trained with no regularization as it provided a superior accuracy. SQ is trained with $\lambda= 10^{-1}$ and $\gamma = 10^{-2}$ and the effective regularization constant is divided by the number of weights in each layer to compensate for the layer size. Validation accuracy for each method is reported in Table~\ref{tab:MNIST_results}, as well as the quantization depth, and the overall compression ratio. We observe that SQ network quantized the first two convolutional layers in 1 bit, and the last fully-connected layers in 2 bits. The accuracy improvement and the compression ratio is marginal for simple task and simple architectures. The effect of smart training becomes more visible for more complex tasks with more layers.  

\begin{table}[h]
\begin{center}
\scalebox{0.9}
{
\begin{tabular}{cccc}
\hline \hline
         \multicolumn{1}{c}{\bf{\bf Method}}
         &\multicolumn{1}{c}{\makecell{\bf Quantization depth \\ \bf per layer (-bits)}} &\multicolumn{1}{c}{\makecell{\bf Compression \\ \bf ratio}} &\multicolumn{1}{c}{\makecell{\bf Accuracy \\ \bf (top-1)}}  \rule{0pt}{4ex} \rule[-1.4ex]{0pt}{0pt} 
\\ \hline \hline 
BC            & 1-1-1-1-1  & 32 & 99.35  \rule{0pt}{4ex} \\
BWN           & 1-1-1-1-1  & 32 & 99.32   \\
TWN           & 2-2-2-2-2  & 16 &  99.38   \\
SQ           & 1-1-2-2-2  & 16.3 & 99.37   \\
FP            &  & 1 & 99.44  \\  \hline
\end{tabular} }
\end{center}
\caption{Smart Quantization (SQ) compared with Binary Connect (BC), Binary Weight Network (BWN), Ternary Weight Network (TWN), and Full Precision (FP) on MNIST dataset. }
\label{tab:MNIST_results}
\end{table}
\subsection{CIFAR10}

\noindent CIFAR10 is an image classification benchmark that contains 32 $\times$ 32 RGB images from ten classes. The dataset is split into 50k training images and 10k testing images. All images are normalized using $\text{mean} = (0.4914, 0.4822, 0.4465)$ and $\text{std} = (0.247,0.243,0.261)$. For the training session, we pad the sides of the images with 4 pixels, then sample a crop of size  $32\times 32$, and flip horizontally at random as our augmentation process.

\begin{table*}[htbp!] 
\begin{center}
\scalebox{.95}
{
\begin{tabular}{ccccc}
\hline \hline  \multicolumn{1}{l}{\bf Architecture}
         &\multicolumn{1}{c}{\bf Method}
         &\multicolumn{1}{c}{\makecell{\bf Quantization depth \\ \bf per layer (-bits)}} &\multicolumn{1}{c}{\makecell{\bf Compression \\ \bf ratio }} &\multicolumn{1}{c}{\makecell{\bf Accuracy \\ \bf (top-1)}} \rule{0pt}{4ex} \rule[-1.4ex]{0pt}{0pt} 
\\ \hline  \hline
& BC            & 1-1-1-1-1-1-1  & 32 & 92.49    \rule{0pt}{3ex} \\
& BWN           & 1-1-1-1-1-1-1  & 32 & 92.42     \\
VGG-7 & TWN     & 2-2-2-2-2-2-2  & 16 & 92.74     \\
& SQ           & 2-1-1-1-2-2-2  & 18.3 & \bf 92.94 \\
& FP            &  & 1 & 93.72  \\
\cline{1-5}
& BC            & 32-1-1-1-1-1-1-1-1-1-1-1-1-32  & 31.5 & 91.92    \rule{0pt}{3ex} \\
& BWN           & 32-1-1-1-1-1-1-1-1-1-1-1-1-32  & 31.5 & 91.85     \\
VGG-16 & TWN    & 32-2-2-2-2-2-2-2-2-2-2-2-2-32  & 15.9 & 92.14     \\
& SQ           & 32-2-1-1-2-2-2-1-1-1-1-1-2-32  & 25.1 & \bf 92.38 \\
& FP            &  & 1 & 92.53  \\ \hline
\end{tabular} }
\end{center}
\caption{Smart Quantization (SQ) compared with Binary Connect (BC), Binary Weight Network (BWN), Ternary Weight Network (TWN), and Full Precision (FP) on CIFAR10 dataset.}
\label{tab:CIFAR_results}
\end{table*}

\begin{figure*} [!htbp]
	\centering
	\begin{subfigure}[t]{0.20\textwidth}
		\includegraphics[width=\textwidth]{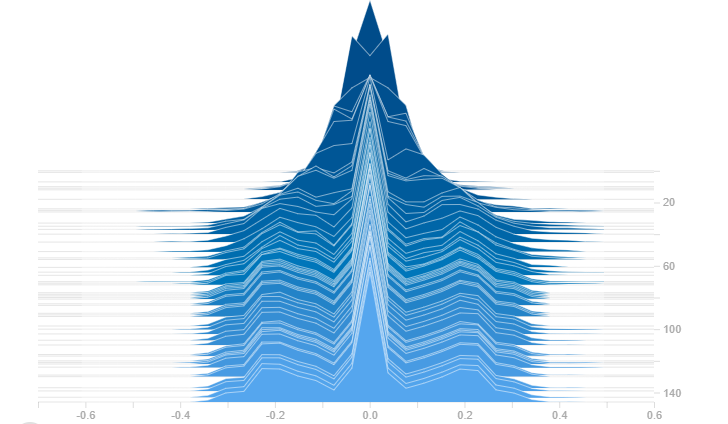}
		\caption{Layer 1, $\beta_1 = 1.56$}
		\label{fig:VGG7_W_dist_a}
	\end{subfigure}
	~ 
	\begin{subfigure}[t]{0.25\textwidth}
		\includegraphics[width=\textwidth]{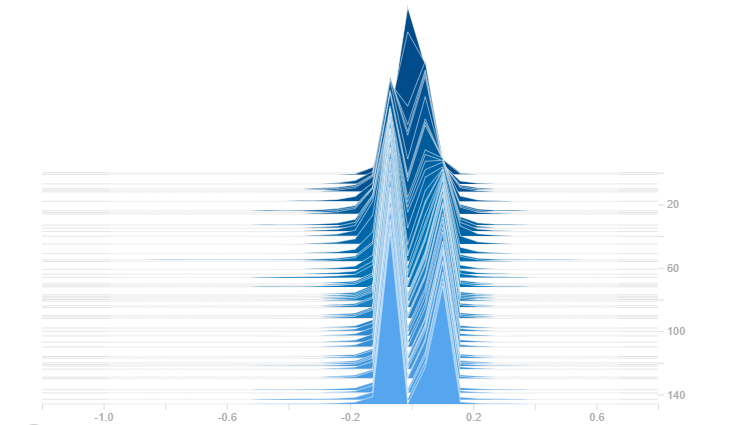}
		\caption{Layer 2, $\beta_2 = 1.57$}
		\label{fig:VGG7_W_dist_b}
	\end{subfigure}
	~ 
	\begin{subfigure}[t]{0.24\textwidth}
		\includegraphics[width=\textwidth]{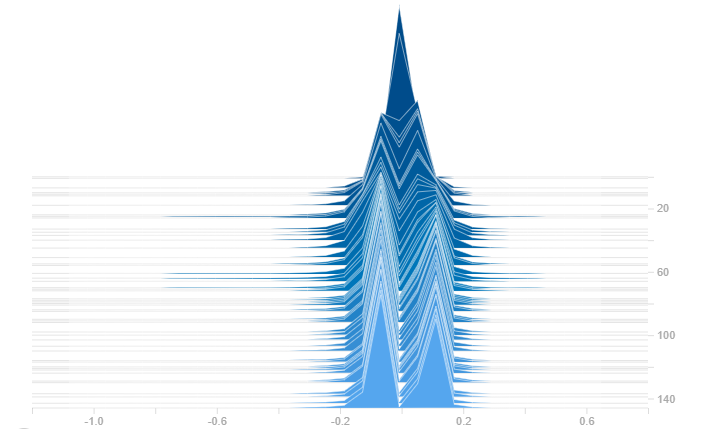}
		\caption{Layer 3, $\beta_3 = 1.57$}
		\label{fig:VGG7_W_dist_c}
	\end{subfigure}
	~
	\begin{subfigure}[t]{0.24\textwidth}
		\includegraphics[width=\textwidth]{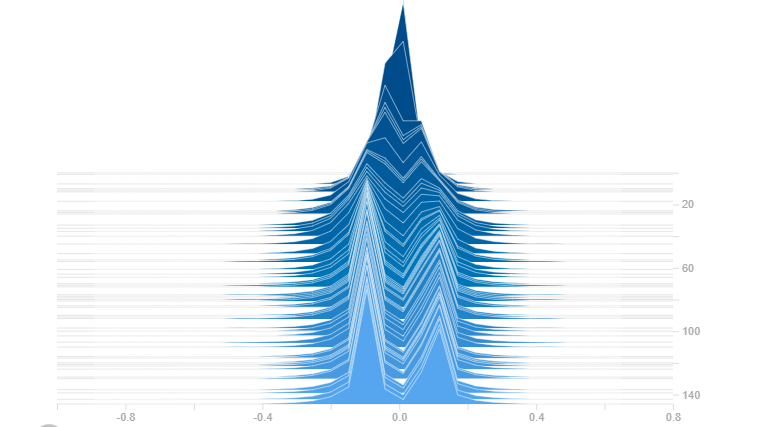}
		\caption{Layer 4, $\beta_4 = 1.57$}
		\label{fig:VGG7_W_dist_d}
	\end{subfigure}
	\\
	\begin{subfigure}[t]{0.25\textwidth}
		\includegraphics[width=\textwidth]{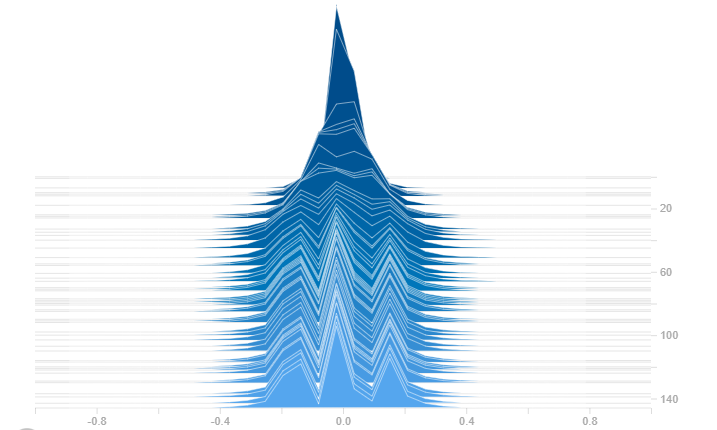}
		\caption{Layer 5, $\beta_5 = 1.00$}
		\label{fig:VGG7_W_dist_e}
	\end{subfigure}
	~ 
	\begin{subfigure}[t]{0.25\textwidth}
		\includegraphics[width=\textwidth]{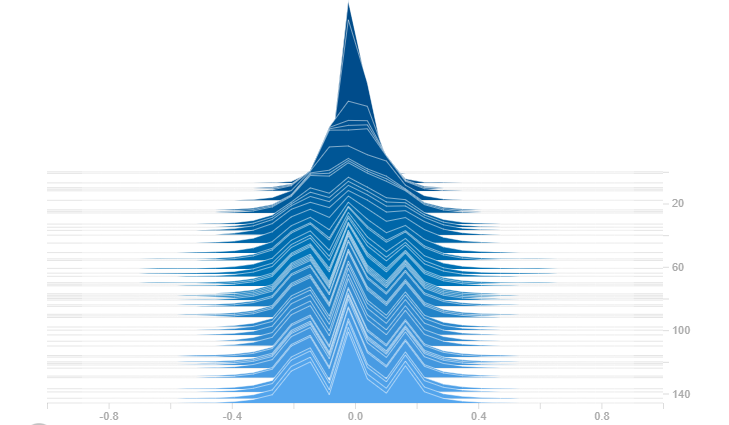}
		\caption{Layer 6, $\beta_6 = 0.86$}
		\label{fig:VGG7_W_dist_g}
	\end{subfigure}
	~
	\begin{subfigure}[t]{0.25\textwidth}
		\includegraphics[width=\textwidth]{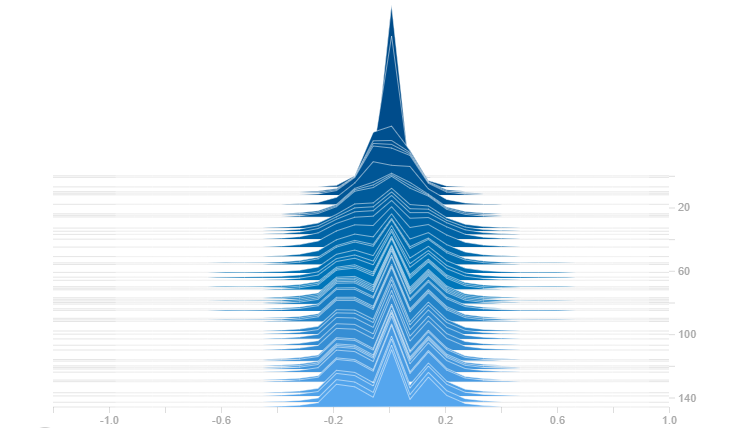}
		\caption{Layer 7, $\beta_7 = 0.98$}
		\label{fig:VGG7_W_dist_h}
	\end{subfigure}
	\caption{Layer-wise weights distribution in VGG-7 for SQ. The weights are pushed to binary when the shape parameter $\beta$ is close to $\frac{\pi}{2} \approx 1.57$.}
	\label{fig:VGG7_W_dist}
\end{figure*}

We use two VGG-like architectures, i) VGG-7 architecture defined in \cite{TWN} in which we apply batch normalization after each layer and use ReLU activations, ii) a standard VGG-16 architecture. We did not quantize the first and the last layers in VGG-16 as the accuracy dropped significantly for all methods.

We train the network for $150$ epochs, using Adam optimizer with the initial learning rate $0.001$ divided by $10$ at epochs $40$ and $80$. The batch size is set to $64$ with $L_2$ weight decay constant $ 10^{-4}$, moreover $\lambda=0.1$, $\gamma= 10^{-3}$ for SQ. Validation accuracy for each method is reported in Table~\ref{tab:CIFAR_results}. SQ beats pure 2 bits network TWN,  even in terms of accuracy. It recommends  three 1 bit layers for VGG-7 and seven 1 bit layers for VGG-16. The compression ratio is significantly higher than a ternary network. The weight distribution of each layers are depicted in Figure~\ref{fig:VGG7_W_dist} for the VGG-7 architecture. Weights are pushed to $\{-\mu,+\mu\}$ or $\{-\mu,0,+\mu\}$ depending on the shape parameter $\beta$.

\section{Conclusions}

We proposed Smart Quantization (SQ), a training method to build a 1 and 2 bits mixed quantized DNN. 
Depth optimization requires training network multiple times which is costly, especially if the network is complex. However, our proposed method successfully combines quantization with different depths, while training the network only once. We focused on layer-wise quantization, since it is more suitable for mixed-precision inference implementation. However, subnetwork, block, filter, or weight mixed quantization is feasible using a similar algorithm. 
 
 SQ makes manual tuning of quantization depth unnecessary. It allows to improve the memory consumption, by automatically quantizing some layers with smaller precision. In some cases, this method even outperforms pure ternary networks in terms of accuracy due to a formal regularization function that shapes trained weights towards mixed-precision.  
It is well-known that some layers are more resilient to aggressive quantization. Our proposed methodology offers a network similar to pure ternary but gives an insight about which layers can be simplified further by adapting to binary quantization. Scaling the method for complex tasks such as ImageNet and with deeper architectures is challenging and requires further research. A similar methodology can be applied to quantize neural network models designed for other tasks related to speech and text.

{\small
\bibliographystyle{ieee_fullname}
\bibliography{egbib}
}

\end{document}

%% file: math_commands.tex
\usepackage{amsmath,amsfonts,bm}

\def\eqref#1{(\ref{#1})}

\def\1{\bm{1}}

\DeclareMathAlphabet{\mathsfit}{\encodingdefault}{\sfdefault}{m}{sl}
\SetMathAlphabet{\mathsfit}{bold}{\encodingdefault}{\sfdefault}{bx}{n}

\DeclareMathOperator*{\argmin}{arg\,min}